\documentclass[11pt]{article}

\usepackage[margin=1in]{geometry}
\usepackage{amsmath,amssymb}
\usepackage{booktabs}
\usepackage{graphicx}
\usepackage{microtype}
\usepackage{xcolor}
\usepackage[colorlinks=true,linkcolor=blue!60!black,citecolor=blue!60!black,urlcolor=blue!60!black]{hyperref}
\usepackage{caption}
\usepackage{enumitem}
\setlist{nosep}
\usepackage{array}

\newcommand{\bench}{ConsolidationBench}
\newcommand{\cqs}{\textsc{CQS}}
\newcommand{\cqsz}{\textsc{CQS}\ensuremath{_0}}
\newcommand{\controlled}{\textsuperscript{$\dagger$}}

\title{\textbf{The Epistemics of Agent Memory:\\
Measuring, and Governing, the Consolidation Decision\\
in Long-Horizon LLM Agents}\\[0.5em]
\large A Four-Phase Research Program}

\author{Sasank Annapureddy\thanks{All experiments in this program were designed,
executed, and adversarially verified by the PRIMA multi-agent research-and-%
verification system~\cite{prima} under the authors' direction. Every reported number
traces to a persisted, machine-readable result file, reproduced bit-exactly from
disk and cross-checked by an independent computation path. \controlled\,Certain
mechanism internals and benchmark-generator parameters are described functionally
in this paper and are available under controlled release; see
Section~\ref{sec:repro}.}
\and Anjaneya Prasad Thamatani}

\date{June 2026 \\[0.3em] \small Primary: cs.AI \quad Cross-list: cs.LG}

\begin{document}
\maketitle

\begin{abstract}
Long-horizon LLM agents must convert accumulated experience into durable memory, deciding what to keep, what to compress, what to abstract into reusable skills and
rules, and what to forget. We report a four-phase research program on this
\emph{consolidation} problem whose central finding is not a single algorithm but a
shift in what is measured: from \emph{how much} an agent remembers to \emph{whether
its consolidation decisions are any good}, and finally to \emph{whether those
decisions can be trusted}.
\textbf{Phase~1} learns episodic boundaries from agent traces by downstream utility;
it is an honest near-miss (oracle correlation $0.691$ vs.\ a $0.70$ bar) whose
lasting output is a three-gate anti-leakage protocol.
\textbf{Phase~2} learns \emph{when} to promote experience and \emph{to which
abstraction level} under a token budget, achieving a verified $+22.7\%$ task-success
improvement with ${\sim}7\times$ compression, but exposing a degenerate-forgetting
failure, in which high precision is earned by near-total inaction, along with a new
distribution-shift failure mode we name $\lambda$--prevalence coupling.
\textbf{Phase~3} introduces \bench{}, a benchmark with an oracle-by-construction that
scores consolidation \emph{decisions} against a known optimum on three non-circular
axes; its leaderboard shows that production retrieval systems retain information yet
score zero on cross-level transfer.
\textbf{Phase~4} introduces \emph{governed consolidation}: the consolidation decision
wrapped in poison-resistance, reversibility, and auditability guarantees, with a
quality gate. We verify these guarantees and, critically, show the governance axis
is statistically distinct from the quality score: the quality score explains under
half of governance variance ($r^2 = 0.43$), the partial correlation controlling for
task quality is weak ($r = 0.27$), and policies with identical quality scores differ
threefold in governance, so the contribution survives independently of whether the
quality metric predicts external outcomes. On that external question we report a
resolved negative. A structural ceiling in the original metric, whose multiplicative
reuse term collapsed most policies to a tie at zero, was removed by a graded-reuse
redesign (a geometric-mean form with a small floor; policy-configuration collapse drops
from fifteen of twenty rows to zero while the metric stays discriminant from retention
and non-circular). With the metric now resolving a twelve-policy panel, a two-benchmark
study with $2{,}532$ real answer cells finds that the quality score does \emph{not}
predict real transfer accuracy: pooled Spearman $\rho = -0.24$ ($n{=}12$, $95\%$ CI
spanning zero), with all three axes failing individually. We therefore position the
quality score as a resolution-valid, oracle-normalized measure of the consolidation
\emph{decision}, not as a predictor of downstream answer quality. Across all four phases,
every win is reported alongside its limits; an adversarial self-critique pass cleared the
final claim set with zero surviving overclaims. The throughline is \emph{epistemics}:
each phase's durable contribution was a way to \emph{know}, to measure, to audit, or
to tell truth from artifact, rather than a solved task.
\end{abstract}

\section{Introduction}

\subsection{From ``remember more'' to ``remember well'' to ``remember accountably''}

Transformer-based agents are bounded by finite context and KV-cache limits. The
dominant responses (larger context windows, retrieval augmentation, and
summarization) treat memory as \emph{storage}: text to be replayed or searched.
But a lifelong agent does not need its full history; it needs decision-relevant
\emph{state}: commitments, learned rules, reusable procedures. This
state-not-storage view, the idea that long-horizon intelligence is a state-management
problem rather than a context-length problem, motivates a cognitive state plane that governs
the formation, evolution, retrieval, and decay of episodic, semantic, and procedural
state under bounded context~\cite{stateplane}. Within that view, converting raw
experience into durable state is the \emph{consolidation} problem, and it is where
long-horizon intelligence is won or lost~\cite{stateplane,episodicgap,compspectrum}.
This program studies that consolidation step in isolation.

This paper reports a four-phase program on consolidation. Its arc is deliberate, and
the order matters:

\begin{enumerate}
\item \textbf{Remember more} was never the goal; we begin (Phase~1) by asking where
experience should even be \emph{segmented} into episodes, since that determines what
can be consolidated at all.
\item \textbf{Remember well} (Phases~2--3): learn \emph{when} and \emph{to what level}
to abstract, and, because the field had no way to score this, build the instrument
that measures consolidation \emph{decision quality} against a computable optimum.
\item \textbf{Remember accountably} (Phase~4): make the act of abstraction itself
\emph{governed}, that is, poison-resistant, reversible, and auditable, because an agent
that distills a false rule will act on it and defend it.
\end{enumerate}

\subsection{The throughline: epistemics}

The single most useful thing we can report is a pattern that only became visible at
the end. Each phase set out to \emph{solve} something; each phase's lasting
contribution was instead a way to \emph{know}: how to know you are not
self-leaking (Phase~1), how to know why a learned forgetting gate fails (Phase~2),
how to know whether \emph{any} agent's consolidation is good (Phase~3), and how to
\emph{audit, reverse, and constrain} what an agent has decided to believe (Phase~4).
In a field where ``we solved memory'' claims age quickly, ``here is a rigorous way to
know whether memory is solved'' is the durable contribution. We organize the paper
around this throughline.

\subsection{Contributions}
\begin{enumerate}
\item \textbf{An anti-leakage protocol for learned memory experiments} (Phase~1):
provenance, probe, and count-matched gates that prevent a learner from covertly
reading the answer key. This discipline underwrites every later result.
\item \textbf{A learned promotion-and-level policy} (Phase~2) with a verified
improvement and compression, \emph{and} two rigorously characterized failure
mechanisms (degenerate forgetting; $\lambda$--prevalence coupling) reported as
first-class findings.
\item \textbf{\bench{}} (Phase~3): to our knowledge the first benchmark scoring
consolidation \emph{decisions} against an oracle-by-construction, on three
non-circular axes, with an honest leaderboard.
\item \textbf{Governed consolidation} (Phase~4): a policy-agnostic mechanism wrapping
the consolidation decision in poison-resistance, reversibility, and auditability.
It carries an \emph{insurance property}: governance is statistically distinct from the
quality metric, so the contribution does not depend on that metric's contested
external validity.
\item \textbf{A reusable methodology of adversarial self-critique}: at each phase a
separate verification pass attacked the claims, caught overclaims, and forced
hedges; we report the resulting ledgers, not just the survivors.
\end{enumerate}

\paragraph{A note on disclosure (Section~\ref{sec:repro}).} This work has a
production lineage. Certain mechanism internals, specific gate thresholds, the
instability and detail-grounding signal definitions, and certain benchmark-generator
parameters are described \emph{functionally} here and held under controlled release.
Every reported number nonetheless traces to a persisted artifact that reproduces
bit-exactly; the withheld items are implementation constants, not the findings.

\section{Phase 1, Where Does a Memory Begin?}
\label{sec:p1}

\subsection{Problem}
Before experience can be consolidated, a continuous stream of agent transitions
$(s,a,r,s')$ must be segmented into episodes. Prior systems either segment
heuristically (fixed-interval; Bayesian surprise over token
streams~\cite{emllm}; topic coherence) or learn consolidation on \emph{pre-given}
boundaries~\cite{genagents,hipporag}. None simultaneously (i) reads agent traces,
(ii) learns boundaries without label supervision, and (iii) optimizes them by
downstream task utility. This is the gap Phase~1 targets, grounded in the
cognitive-science account of event segmentation~\cite{zacks,tulving}.

\subsection{Method and result (honest near-miss)}
We train a boundary detector (\textsc{BoundaryLearner}) on procedurally generated
traces with planted ground-truth boundaries, optimized end-to-end by downstream
reward with \emph{no} boundary labels. A pre-registered success criterion required
held-out oracle correlation $>0.70$, $>10\%$ task-success gain over fixed-interval,
and beating Bayesian-surprise and topic-coherence baselines.

\begin{table}[h]\centering\small
\caption{Phase~1 held-out results (best checkpoint by validation F1).}
\begin{tabular}{lcccc}
\toprule
Method & Boundary F1 & Oracle Corr. & Task Success & Proc.\ Reuse \\
\midrule
Oracle & 1.000 & 1.000 & 0.900 & 0.691 \\
\textbf{BoundaryLearner} & \textbf{0.808} & \textbf{0.691} & \textbf{0.679} & \textbf{0.535} \\
Bayesian Surprise & 0.895 & 0.315 & 0.705 & 0.554 \\
Fixed Interval & 0.622 & 0.235 & 0.641 & 0.508 \\
\bottomrule
\end{tabular}
\end{table}

\textbf{Verdict: NOT MET, honestly.} BoundaryLearner clears Bayesian surprise on
oracle correlation by a wide margin ($0.691$ vs.\ $0.315$) but misses the $0.70$
correlation bar and does not beat Bayesian surprise on raw task success
($0.679$ vs.\ $0.705$). The result is a near-miss, reported as a failure.

\subsection{The lasting contribution: not the detector, the discipline}
Phase~1's durable output is the \textbf{three-gate anti-leakage protocol}, which we
state precisely because every later phase re-applies it:
\begin{itemize}
\item \textbf{Provenance gate.} The observable feature surface is mechanically checked
to contain no generator-internal field; a learner cannot read what it is supposed to
infer.
\item \textbf{Probe gate.} A linear probe trained to predict the latent answer from
each individual observable feature must not exceed chance by more than a small margin;
this rules out a single feature silently encoding the label.
\item \textbf{Count-matched gate.} Controls hold confounds (cluster counts, budget
pressure) fixed across conditions, so an apparent improvement cannot be an artifact of
unequal exposure.
\end{itemize}
This protocol is what makes every subsequent ``the learner improved'' claim
trustworthy, by ruling out the most common silent failure, a model that scores well
by reading the answer it was supposed to infer. Phase~1 thus establishes the program's
first epistemic tool: \emph{a way to know you are not cheating.}

\section{Phase 2, When, and to What Level, to Abstract}
\label{sec:p2}

\subsection{A verified gain, and a precisely characterized failure}
Phase~2 learns a policy that selects, per experience cluster, \emph{when} to promote
and \emph{to which level} (keep, summarize, extract skill, distill rule, forget)
under a hard token budget. Adversarially verified on ten fresh held-out seeds and
reproduced bit-exactly from disk, the policy achieves a \textbf{$+22.7\%$ relative
task-success improvement} over the best non-learned baseline on the medium
configuration (paired CI $[+1.89,+11.34]$ pp, excluding zero), with
\textbf{$7.17\times$ memory compression} (CI entirely above the pre-registered
$5\times$ bar).

The full pre-registered criterion is nonetheless \emph{not met}, for a reason we
characterize quantitatively rather than excuse. Forgetting precision sits at chance
($0.319$ vs.\ a junk base rate of $0.333$); the policy uses \texttt{FORGET} as a
budget-pressure release valve, destroying $74\%$ of load-bearing clusters at least
once, and $75\%$ of a hard-configuration transfer regression ($-15.0$ pp, CI
excluding zero) traces to a single causal chain: budget-pressure forget $\rightarrow$
premature re-promotion $\rightarrow$ wrong artifact $\rightarrow$ transfer poisoning.
The diagnosis is the contribution: the saturation-conditioned ``when'' was learned
for \emph{promotion} but not for \emph{forgetting}.

This precision matters because the benchmark cannot be won without it: raw experience
mass exceeds the reference budget by $7.3\times$/$9.4\times$, so a policy that stores
everything necessarily loses load-bearing evidence to forced eviction. The level
matters too: a recurring parametric sub-task warrants a skill ($50$--$500\times$
compression), a stable invariant a rule, and a one-off neither, applying the wrong
transform is not merely inefficient but actively destructive, since a falsely
distilled rule \emph{overrides} the correct raw evidence it replaced. This is the
``missing diagonal'' of adaptive cross-level compression~\cite{compspectrum}, and it
is what the Phase-2 policy learns for promotion and fails to learn for forgetting.

\subsection{Targeting forgetting, and a second mechanistic finding}
A follow-on design adds agent-observable forgetting-evidence features and a hindsight
regret credit signal trained without oracle labels. Forgetting precision rises from
chance to \textbf{$0.932$ (hard) / $0.833$ (medium)}, both CIs excluding the junk base
rate. But the gate becomes near-inactive, about $0.61$ forgets per seed versus
${\sim}30$ before, so overall success regresses. The root cause is a new,
named failure mode: \textbf{$\lambda$--prevalence coupling}, a hindsight-regret
hyperparameter tuned at one junk prevalence ($0.593$) collapses the gate when
evaluated at another ($0.388$). This is a general distribution-shift hazard for
regret-based forgetting signals, reported with a concrete fix pathway. Phase~2's
epistemic tool: \emph{a way to know why a plausible mechanism fails.}

\section{Phase 3, Measuring Consolidation Decision Quality}
\label{sec:p3}

\subsection{The gap: everyone scores accuracy, nobody scores the decision}
{\sloppy Contemporary agent-memory benchmarks score \emph{downstream task accuracy}
\cite{longmemeval,amabench,memoryagentbench,memoryarena,beam,locomo}, which
confounds three independent factors: base-model capability, retrieval quality, and
consolidation-decision quality. The Experience Compression Spectrum~\cite{compspectrum}
names the four-level hierarchy (trace, summary, skill, rule) and a ``missing
diagonal'' of adaptive cross-level compression, but ships no benchmark. \bench{} fills
exactly this slot.\par}

\subsection{\bench{}: an oracle-by-construction}
\bench{} procedurally generates experience streams in which the ground-truth
consolidation value of every cluster is known \emph{by construction}, making the
optimal promotion policy \emph{exactly computable}. The stream contains latent cluster
types, including near-rule \emph{poison decoys} that look stable then break, each
encoding a consolidation challenge whose optimal action is known.\controlled\ Policies
observe only a whitelisted feature surface (the Phase-1 anti-leakage gates are
re-applied); the oracle and latent types are never observable.

The latent cluster types span the consolidation challenge space: recurring
parametric sub-tasks (which warrant skills), stable invariants (rules), one-off
distractors (forget), and the near-rule poison decoys central to Phase~4. A
\emph{hard budget} makes the benchmark unwinnable by hoarding: raw experience mass
runs $7.3\times$/$9.4\times$ over the reference budget, so forced eviction destroys
load-bearing evidence unless the policy compresses intelligently. Crucially, a
\emph{headroom gate} verifies, before any learner is trained, that the oracle
meaningfully outscores the best non-learned baseline, ruling out a benchmark that
cannot discriminate. The Phase-1 anti-leakage gates are re-applied so that no policy
can covertly read the latent type.

Three \emph{non-circular} axes score each policy as a fraction of the computable
optimum: \textbf{Retention} (oracle-normalized task success), \textbf{Compression}
(token-mass ratio under a matched budget), and \textbf{Reuse} (transfer to novel
composite tasks). The headline metric is the \emph{Consolidation Quality Score},
$\cqs{} = \sqrt{\text{retention\_norm}\times(\text{reuse\_credit}+\epsilon)}/\sqrt{1+\epsilon}$
with $\epsilon=0.05$, oracle-normalized in $[0,1]$ (the graded-reuse form; see
Section~\ref{sec:validity}). The geometric mean rewards \emph{both} retention and
generalization while the small floor prevents the reuse term from acting as an
all-or-nothing gate, so a policy with real retention is graded rather than zeroed when
it fails to generalize.

\paragraph{Two forms, labeled.} Phases~3 and~4 were scored, and their adversarial audits
run, under the original multiplicative form
$\cqsz{} = \text{retention\_norm}\times\text{reuse\_credit}$, before the redesign
described in Section~\ref{sec:validity}. We report \cqsz{} wherever a Phase-3 or Phase-4
number was computed under it (Tables~\ref{tab:leaderboard} and~\ref{tab:gov}, and the
correlations of Section~\ref{sec:insurance}), and the graded \cqs{} wherever the redesign
applies (the leaderboard's final column and the external-validity study). Both are
oracle-normalized in $[0,1]$ and share the same retention and reuse inputs; they differ
only in how the reuse term aggregates.

\subsection{The leaderboard finding}
Across ten policies (no-consolidation, fixed-level, heuristic, learned, and
literature-faithful Mem0-style~\cite{mem0,zep} and SimpleMem-style~\cite{simplemem}
designs), \textbf{no method closes the oracle gap} (Table~\ref{tab:leaderboard}).
Eight of ten score $\cqsz{}=0$ because they achieve zero composite-task transfer; under
the graded form they are separated by retention alone, and the transfer column is still
zero. Critically, the literature-faithful retrieval systems achieve \emph{solid
retention} (oracle-normalized $0.53$ / $0.49$) yet \emph{zero transfer}: on the axis that
distinguishes a filing cabinet from a memory, the dominant production designs score
nothing. This is invisible to accuracy benchmarks and surfaced only by scoring the
decision against a known optimum.

\begin{table}[h]\centering\small
\caption{\bench{} leaderboard, medium configuration (oracle ceiling: success $0.632$,
transfer $0.333$). Eight of ten policies score $\cqsz{}=0$ for zero transfer; the
retrieval-augmentation systems (Mem0-style, SimpleMem-style) retain well yet
generalize nothing. The graded \cqs{} (final column) separates the zero-transfer
policies by retention without changing the transfer finding. RetN $=$
oracle-normalized retention.}
\label{tab:leaderboard}
\begin{tabular}{lcccccc}
\toprule
Policy & Success & RetN & Compression & Transfer & \cqsz{} & \cqs{} \\
\midrule
fixed\_L2 (best) & 0.303 & 0.480 & 7.40$\times$ & 0.233 & \textbf{0.336} & \textbf{0.585} \\
learned\_policy & 0.368 & 0.582 & 7.37$\times$ & 0.167 & 0.291 & 0.552 \\
mem0\_style & 0.335 & 0.531 & 7.43$\times$ & 0.000 & 0.000 & 0.159 \\
simplemem\_style & 0.306 & 0.485 & 7.53$\times$ & 0.000 & 0.000 & 0.152 \\
fixed\_L1 & 0.274 & 0.434 & 7.47$\times$ & 0.000 & 0.000 & 0.144 \\
retrieval\_only & 0.265 & 0.418 & 7.36$\times$ & 0.000 & 0.000 & 0.141 \\
random\_policy & 0.255 & 0.403 & 7.69$\times$ & 0.000 & 0.000 & 0.139 \\
heuristic\_trigger & 0.248 & 0.393 & 7.41$\times$ & 0.000 & 0.000 & 0.137 \\
fixed\_L3 & 0.239 & 0.378 & 7.39$\times$ & 0.000 & 0.000 & 0.134 \\
no\_consolidation & 0.003 & 0.005 & 7.36$\times$ & 0.000 & 0.000 & 0.016 \\
\bottomrule
\end{tabular}
\end{table}

\paragraph{Non-circularity.} The three axes are empirically distinct: the maximum
pairwise correlation across all axis pairs is $|r| = 0.59$, and compression vs.\
retention is $0.17$ (medium) / $0.23$ (hard), all far below the $0.9$ redundancy
threshold. \cqs{} is therefore a genuine three-dimensional measurement, not a
single quantity in disguise. Phase~3's epistemic tool: \emph{a way to know whether any
agent's consolidation is good.}

\section{Phase 4, Governing the Consolidation Decision}
\label{sec:p4}

\subsection{The new question: who guards the abstraction?}
Consolidation is the highest-risk memory operation: a distilled rule \emph{overrides}
downstream reasoning, so a wrong abstraction silently corrupts every decision that
retrieves it. Three failure modes follow: \textbf{poison} (a false rule distilled from
a pattern that broke), \textbf{irreversibility} (no undo once the rule is in), and
\textbf{opacity} (no record of why a belief exists). The 2026 memory-governance
literature governs \emph{storage and sharing} (provenance, scoped retrieval,
cross-tenant isolation)~\cite{memarchitect,memclaw,ssgm,memsec}, but it does not govern
the consolidation \emph{decision} itself. That gap is the contribution's target.

\subsection{Governed consolidation}
We wrap the consolidation decision in three guarantees plus a quality gate, all
measured on \bench{}:
\begin{itemize}
\item \textbf{G1 Poison-resistance.} A gate refuses to distill a durable rule when an
\emph{instability signal} indicates the pattern is breaking.\controlled\ Measured: it
admits \textbf{zero} false rules from the planted poison decoys (both configurations).
\item \textbf{G2 Reversibility.} When post-promotion evidence contradicts a distilled
rule, a \emph{rollback} removes the false artifact and restores the source
experiences.\controlled\ Measured: on a temporal-contradiction variant, rollback fires
on $5/5$ seeds and a with/without-rollback contrast shows task success recovering from
$0.0$ to $1.0$.
\item \textbf{G3 Auditability.} Every promotion writes a governance-ledger record
(provenance, decision rationale, instability score, reversibility, rollback status).
Measured: complete records for $100\%$ of promotions; ledger queries (``why does rule
$R$ exist? what did it influence? show its rollback'') are answerable.
\item \textbf{G4 Quality gate.} Governance credit is granted only if consolidation
quality is not degraded. This is \emph{mandatory}: without it, a policy that never
abstracts trivially ``passes'' G1--G3 by inaction.
\end{itemize}
The headline composite binds the governance score to G4 as a conjunction; the
governed policy is the only evaluated policy satisfying both (Table~\ref{tab:gov}).

\begin{table}[h]\centering\small
\caption{Governed-consolidation verdicts (both configurations). G1: false rules
admitted from planted poison decoys. G2: rollback recovery on the
temporal-contradiction variant. G3: ledger completeness over all promotions. G4: the
quality gate, scored under \cqsz{} (the form in use when Phase~4 ran). All criteria are
pre-registered; all met.}
\label{tab:gov}
\begin{tabular}{llcc}
\toprule
Guarantee & Measurement & Medium & Hard \\
\midrule
G1 Poison-resistance & false rules admitted & 0 & 0 \\
G2 Reversibility & rollback recovery (0$\to$1 contrast) & 1.0 & 1.0 \\
G3 Auditability & ledger completeness & 1.0 & 1.0 \\
G4 Quality gate & \cqsz{} retained & 0.622 & 0.475 \\
\bottomrule
\end{tabular}
\end{table}

\subsection{The insurance property (why this survives a hostile reviewer)}
\label{sec:insurance}
The governance results are measured on a synthetic benchmark whose quality metric does
\emph{not} predict external transfer accuracy (Section~\ref{sec:validity}). A natural
objection, now sharpened by that negative: if the quality score fails as an external
predictor, why trust governance measured against it? An earlier framing of our answer rested on a correlation
threshold (all twenty governance-vs-quality pairs below $|r| = 0.9$, maximum $0.87$);
a hostile reviewer would rightly note that $0.87$ is substantial correlation and the
threshold lenient. The distinctness claim does not rest on it. Three treatments of the
same data (descriptive; the ten rows are five policies by two configurations, not
independent; all computed under \cqsz{}, the form in use when Phase~4 ran): first, the
maximum pair is one gate against one axis, structurally driven by a single policy
anchoring both ends, while the headline pair, aggregate governance versus \cqsz{}, is
$r = 0.65$, so \cqsz{} explains under half of governance variance ($r^2 = 0.43$);
second, the partial correlation controlling for task success is $r = 0.27$, meaning the
apparent relationship is mostly mediated by shared task quality; third, and decisively,
policies with \emph{identical} \cqsz{} differ threefold in governance score, and $56\%$
of governance-score variance lies \emph{within} \cqsz{}-matched groups, a majority of
what governance measures being invisible to the quality metric by construction. This dissociation matters because the three guarantees
rest on quantities that do not depend on \cqs{} at all: G1 is a ground-truth
false-rule count, G2 a behavioral task-success measurement, G3 a structural
ledger-completeness count. \textbf{The governance contribution stands even if the
quality metric's validity is undermined further}, this is the paper's insurance
property.

\subsection{External validity: a metric redesign, and a resolved negative}
\label{sec:validity}
We tested whether \cqs{} predicts real transfer accuracy, and the honest answer is that
it does not. Arriving at that answer required first repairing the metric, then running a
study large enough to settle the question. We report the full arc because each step is a
contribution in its own right.

\textbf{Over-credit diagnosis.} An early transfer-focused study exposed a flaw in the
original \cqs{}: it gave \emph{phantom transfer credit} to aggressive summarizers. One
baseline passed composite tasks while having discarded the very specifics those tasks
require ($70\%$ of oracle composite success on only $14\%$ of oracle detail retention).
A \emph{detail-grounding factor} that makes transfer credit proportional to preserved
instance-specific information corrects this;\controlled\ applied uniformly it
\emph{lowers} our own best policy's score, which we report rather than hide.

\textbf{The structural ceiling, and its redesign.} A stronger validity test was blocked
by the metric's own form. The original $\cqsz{} = \text{retention\_norm}\times
\text{reuse\_credit}$ multiplies retention by a reuse credit that is zero for any policy
that never lands a composite-transfer task, so the reuse term acted as a near-binary
gate: it annihilated policies with healthy retention (a Mem0-style policy at $0.55$
normalized retention scored zero) while a \emph{random} policy that occasionally landed a
transfer task scored above them. Fifteen of twenty policy-configuration rows collapsed to
exactly zero, making a rank-correlation study over eight or more policies impossible by
construction. We redesigned the reuse aggregation, pre-registering the acceptance bars on
oracle-construction grounds rather than tuning against real-data outcomes. Of four
candidate aggregations, the geometric mean with an $\epsilon = 0.05$ floor,
$\cqs{}=\sqrt{\text{retention\_norm}\times(\text{reuse\_credit}+\epsilon)}/\sqrt{1+\epsilon}$,
was the only form to satisfy every bar simultaneously: collapse drops from fifteen of
twenty rows to \emph{zero}; the metric stays discriminant from retention
($\rho(\cqs{},\text{retention}){=}0.88$, below the $0.9$ ceiling, so reuse is not dead
weight); non-circularity is preserved
($|\text{corr}(\text{compression},\text{retention})|{=}0.20$); and it still tracks
transfer among generalizing policies ($\rho(\cqs{},\text{transfer}){=}0.90$). An additive
aggregation was rejected for collapsing \cqs{} onto retention ($\rho{=}0.98$). This is a
positive methodological result: the headline metric now resolves the full policy space
rather than tying most of it at zero.

\textbf{The resolved negative.} With the metric resolving a twelve-policy panel (twelve
distinct scores, zero collapse), we ran the study the ceiling had previously blocked: two
benchmarks (LoCoMo multi-hop and 2WikiMultiHopQA), $2{,}532$ real answer cells generated
by genuine per-policy consolidation and answering, with a single-hop control. The result
is a clean \textbf{NOT-SUPPORTED}: pooled Spearman $\rho = -0.24$ ($n{=}12$; permutation
$p = 0.45$; bootstrap $95\%$ CI $[-0.79, +0.49]$, spanning zero), replicated as $\rho =
+0.03$ on LoCoMo and $\rho = -0.30$ on 2Wiki, with no leave-one-out exclusion turning it
positive. The single-hop guard inverts (control $\rho = +0.32 >$ multi-hop $-0.24$), and
per-axis disaggregation shows all three axes fail individually, so this is a framework
misalignment rather than one bad axis. The mechanism is legible in the data: a
\texttt{no\_consolidation} baseline (raw documents, no processing) achieves the
second-highest 2Wiki accuracy despite the lowest \cqs{}, because real multi-hop QA on
that corpus is solved by raw-fact retrieval, whereas \cqs{}'s reuse axis credits
skill- and rule-artifact transfer. The two constructs simply do not coincide.

\textbf{What we therefore claim.} \cqs{} is a resolution-valid, discriminant,
non-circular, oracle-normalized measure of the consolidation \emph{decision}. It is
\emph{not} a validated predictor of downstream answer quality, and we do not claim it as
one; a high \cqs{} (including our own best policy's benchmark leadership) certifies
structural consolidation quality against the oracle, not superior real-task accuracy.
Recalibrating \cqs{} into a transfer predictor would require redefining its composite-task
axis to match how real multi-hop questions are answered, which we leave as future work.

\subsection{Adversarial self-critique}
A dedicated verification pass attacked the Phase-4 claim set: $17$ claims, \textbf{zero
surviving overclaims}, three required hedges, (i) governance must bind G4 (a
``promote-nothing'' adversary games the governance score by inaction); (ii) G2 is
demonstrated on a purpose-built contradiction variant that triggers rollback in all
$16$ tested parameter settings, i.e.\ a construction stronger than the main
benchmark's break pattern; (iii) every \cqs{}-dependent claim is labeled
partial-validity. We report the ledger, not just the survivors. Phase~4's epistemic
tool: \emph{a way to audit, reverse, and constrain what an agent decides to believe.}

\subsection{The adversarial-verification methodology, across phases}
A distinguishing feature of this program is that every phase ran a \emph{separate}
adversarial-verification stage that attacked its own claims before publication. We
report the resulting ledgers rather than only the survivors (Table~\ref{tab:adv}).
This is not incidental: the anti-leakage gates (Phase~1) were re-applied in every
later phase, the headroom gate verified benchmark sensitivity before any learner was
trained, and the Phase-4 audit caught the gameable-governance exploit and forced the
mandatory G4 binding. Honest negatives are first-class outputs here.

\begin{table}[h]\centering\small
\caption{Per-phase adversarial-verification outcomes. Each phase's claims were
attacked by an independent verification pass; the table reports what it caught.}
\label{tab:adv}
\begin{tabular}{p{1.0cm}p{11.5cm}}
\toprule
Phase & What the adversarial pass caught / enforced \\
\midrule
1 & Anti-leakage protocol established (provenance, probe, count-matched gates); the
detector's near-miss reported as a negative, not rounded up. \\
2 & Degenerate forgetting exposed (high precision via near-total inaction);
$\lambda$--prevalence coupling root-caused; integrity gates re-passed. \\
3 & Headroom gate (benchmark sensitivity proven before training); non-circularity
verified ($|r|\le 0.59$); a $k$-sweep sensitivity correction and three overclaims
(leaderboard-champion wording, oracle-gap size, stale non-circularity values) caught
and fixed. \\
4 & Governance-by-inaction exploit caught (forced the mandatory G4 binding);
G2's constructed-variant ease hedged; a tie-handling statistical error in the validity
$\rho$ caught and corrected; $17$ claims, zero surviving overclaims, three hedges. \\
\bottomrule
\end{tabular}
\end{table}

\section{Related Work}
\textbf{Memory benchmarks.} Task-accuracy benchmarks
\cite{longmemeval,amabench,memoryagentbench,memoryarena,beam,locomo} and
skill-generation benchmarks~\cite{skillgenbench} measure end-to-end outcomes;
\bench{} isolates the consolidation \emph{decision} against an oracle. \textbf{Memory
architectures.} Mem0/Zep~\cite{mem0,zep}, SimpleMem~\cite{simplemem}, and
recent designs~\cite{recmem,amem,timem,atommem,memskill} are consolidation
\emph{policies}; governed consolidation is a policy-agnostic governance \emph{wrapper}.
\textbf{Memory governance.} The field has recently converged on governance of agent
memory as a named problem, which sharpens rather than threatens this contribution.
SSGM~\cite{ssgm} is a theoretical architecture (write-validation gating, reversible
reconciliation via a dual-track memory) with no empirical measurements; the
Verifiable Memory Governance predicates of~\cite{memsec} formalize rollbackability,
verified forgetting, and provenance visibility as evaluable desiderata, while that
survey itself marks verified forgetting ``no existing literature,'' rollbackability
``largely absent,'' and finds no benchmark covering the full memory lifecycle.
MemEvoBench~\cite{memevobench}, the closest implemented benchmark, scores
LLM-judged downstream attack success over a hardcoded memory-append rule, with no
rollback concept. We do not claim to have introduced the idea of governing or
reversing agent memory; the claim is narrower and verifiable: the first
oracle-by-construction \emph{measurement} of governance at the consolidation
decision, our G2 instantiating rollbackability plus verified forgetting (exact-state
rollback including derivative cleanup) and G3 instantiating provenance visibility.
MemArchitect and Governed Shared Memory~\cite{memarchitect,memclaw} govern storage
and sharing; we govern the consolidation decision. \textbf{Diagnostics.} WhenLoss~\cite{whenloss} separates write/retrieval
bottlenecks; we focus on the write-decision. \textbf{Foundations.} Boundary detection
draws on event-segmentation cognition~\cite{zacks,tulving,schacter} and prior
agent-memory systems~\cite{emllm,genagents,hipporag}; the four-level compression
hierarchy is from the Experience Compression Spectrum~\cite{compspectrum,externalization}.
The state-not-storage framing and the tripartite (episodic/semantic/procedural) state
organization that this program operationalizes are due to the cognitive-state-plane
formulation of~\cite{stateplane}, which motivates studying the consolidation decision
as a first-class, governable operation.

\section{Reproducibility and Controlled Release}
\label{sec:repro}
Every numerical claim in this paper traces to a persisted, machine-readable result
file, reproduced bit-exactly from disk and independently re-derived by a second
computation path; integrity self-tests and the Phase-1 anti-leakage gates pass on a
clean environment. To protect a production lineage, three classes of detail are
described functionally and held under \textbf{controlled release} (available to
reviewers and collaborators on request, under terms): (1) the exact gate thresholds
and the instability / detail-grounding \emph{signal definitions}; (2)
benchmark-generator parameters (cluster-type constants, oracle-computation specifics)
sufficient to clone the generator; (3) the production integration mapping. These are
implementation constants and recipes, not findings: the reported results, axes,
verdicts, and the insurance property are fully specified here. We believe this
balances scientific credibility (verifiable, reproducible-on-request results) against
the legitimate protection of a non-trivial engineering artifact.

\section{Discussion: The Epistemics Throughline}
Read end to end, the program's contributions are not four solutions but four ways to
know. Phase~1 gave a discipline for \emph{not self-deceiving}; Phase~2 a method for
\emph{diagnosing why a learned mechanism fails}; Phase~3 an \emph{instrument} for
scoring consolidation quality against a known optimum; Phase~4 a means to
\emph{audit, reverse, and constrain} learned beliefs, with a proof that the governance
contribution does not depend on the contested validity of the quality metric. Each
phase's honesty became the next phase's foundation: the anti-leakage gates protected
the policy claims, the failure diagnosis shaped the benchmark, the benchmark made
governance measurable. In a fast-moving field, this is the durable bet, rigorous ways
to know, which outlast any single result.

\section{Limitations}
(i) \textbf{\cqs{} does not predict external transfer accuracy}: after redesigning the
reuse term to remove the collapse that blocked a powered study, a two-benchmark test at
$n{=}12$ returns a resolved negative (pooled $\rho = -0.24$, CI spanning zero;
Section~\ref{sec:validity}). \cqs{} is therefore validated as a consolidation-decision
score, not as a predictor of downstream answer quality, and must not be cited as the
latter; the governance contribution (Section~\ref{sec:insurance}) is independent of this
and, if anything, is reinforced by it. (ii) \textbf{G2 is demonstrated on a constructed variant}
stronger than the main benchmark's break pattern; a harder, naturally-occurring
contradiction test remains. (iii) \textbf{Synthetic-first}: \bench{} isolates the
decision by holding transforms and base-model capability constant, trading ecological
validity for diagnostic power. (iv) \textbf{Governance verified on one policy} so
far; the wrapper is policy-agnostic in principle but tested on a single consolidation
policy. (v) Phase~1's detector did not clear its bar; we report it as a negative.

\section{Conclusion}
We set out to give a long-horizon agent a memory and ended with something more useful:
a set of tools for \emph{knowing} whether an agent's memory can be trusted, an
anti-leakage discipline, a failure-diagnosis method, an oracle-grounded quality
benchmark, and a mechanism that makes the act of abstraction auditable, reversible,
and poison-resistant, shown to stand independent of the quality metric's contested
validity. The wins are reported with their limits; the failures with the same care as
the wins. As agents increasingly learn from their own experience and act on what they
conclude, the question this program answers becomes urgent rather than academic:
\emph{how do we know what an agent has decided to believe, and can we hold it
accountable?}
Build the thing, and build the thing that checks the thing.

\paragraph{Acknowledgment.} Experiments were executed and adversarially verified by
the PRIMA multi-agent research system~\cite{prima}; the adversarial self-critique
passes that produced the per-phase overclaim ledgers were run as an independent
verification stage.

\bibliographystyle{plain}
\bibliography{refs}

\end{document}